# SWARM-BASED SPATIAL SORTING

MARTYN AMOS

*Department of Computing and Mathematics, Manchester Metropolitan University, Chester Street, Manchester, M1 5GD, United Kingdom*
M.Amos@mmu.ac.uk
http://www.martynamos.com

OLIVER DON

*School of Engineering, Computer Science and Mathematics, University of Exeter, North Park Road, Exeter, EX4 4QF, United Kingdom*
oliver.don@symbian.com.

**Purpose** – To present an algorithm for spatially sorting objects into an annular structure.
**Design/Methodology/Approach** – A swarm-based model that requires only stochastic agent behaviour coupled with a pheromone-inspired "attraction-repulsion" mechanism.
**Findings** – The algorithm consistently generates high-quality annular structures, and is particularly powerful in situations where the initial configuration of objects is similar to those observed in nature.
**Research limitations/implications** – Experimental evidence supports previous theoretical arguments about the nature and mechanism of spatial sorting by insects.
**Practical implications** – The algorithm may find applications in distributed robotics.
**Originality/value** – The model offers a powerful minimal algorithmic framework, and also sheds further light on the nature of attraction-repulsion algorithms and underlying natural processes.

## 1. Introduction

The ability of social insects to collectively solve problems has been well-studied and documented (Camazine *et al.*, 2001). The behaviour of foraging ants, for example, has been abstracted to provide algorithmic solutions that are robust, distributed, and flexible (Dorigo *et al.*, 1999; Dorigo and Stützle, 2004). The particular behaviour that we will focus on is the clustering or sorting of ant corpses or larvae (Deneubourg *et al.*, 1991). Abstract models of these behaviours have been successfully applied to, amongst other problems, numerical data analysis, data mining, and graph partitioning (Handl *et al.*, 2003). In this paper, we focus on the task of *brood sorting*. This behaviour, when observed in *Temnothorax unifasciatus*[1] (Franks and Sendova Franks, 1992), leads to the formation of a single cluster of offspring made up of concentric rings of brood items, with the youngest items (eggs and micro-larvae) being tightly packed at the centre, and successively larger larvae arranged in increasingly wider-spaced bands moving out from the centre of the cluster. Models of this behaviour have yielded new algorithmic solutions to the problem of annular sorting (Scholes *et al.*, 2004; Wilson *et al.*, 2004; Hartmann, 2005; Vik, 2005; Scheidler *et al.*, 2006).

In this paper we present a novel algorithm inspired by an intriguing hypothesis by Franks and Sendova-Franks concerning the biological mechanisms underlying annular sorting. In their article, the authors state that "The mechanism that the ants use to re-create these brood patterns when they move to a new nest is not fully known. Part of the mechanism may involve conditional probabilities of picking up and putting down each item which depend on each item's neighbours ... The mechanisms that set the distance to an item's neighbour are unknown. They may be pheromones that the brood produce and which tend to diffuse over rather predictable distances ..."(Franks and Sendova Franks, 1992)

---

[1]Until recently, this species was known as *Leptothorax unifasciatus*.

We should note that the purpose of this paper is *not* to investigate the actual biological phenomenon in question; we simply use it as inspiration for developing a new algorithmic technique. We have constructed a corresponding minimal model, using only stochastic agent behaviour and fixed "repellents" and "attractants". The general notion of "attraction-repulsion" as a spatial sorting mechanism has been well-studied by biologists, and is applicable to organisms as diverse as ants, fish and birds (Parrish and Hamner, 1997; Okubo, 2001). Our model gives rise to the emergence of annular clusters of objects in simulated colonies. Moreover, it is competitive with existing solutions for simple sorting, and has the additional properties of being able to deal with both objects of non-uniform physical size and pre-sorted piles of items.

In Section 2 we present the background to the problem, before describing our model in Section 3. In Section 4 we describe in detail the metrics for assessing the quality of solutions generated, and in Section 5 we present and discuss the results of experimental investigations (including extended parametric and convergence analyses). We conclude in Section 6 with a discussion of the implications of our findings. This article is an extended version of work first presented in (Amos and Don, 2007).

## 2. Annular Sorting

(Franks and Sendova Franks, 1992) carried out an observational biological study of the brood sorting behaviour of *Temnothorax* ants. The nesting behaviour of this species made them particularly suitable for study as they nest in single clusters in flat rock crevices, a situation that is easy to replicate and monitor in a laboratory. Photographs were taken of the ants' brood cluster and individual items were classified, before a tessellation was applied to determine density and distance from the cluster centroid. This study showed that *Temnothorax* ants placed smaller brood items at the centre of the cluster with a greater density, forming an annular arrangement. This process reasserted itself when the ants were forced to migrate to a new nesting site, and proved to be ongoing. This structure is illustrated in Figure 1, where three different types of brood item are arranged in roughly-sorted concentric rings.

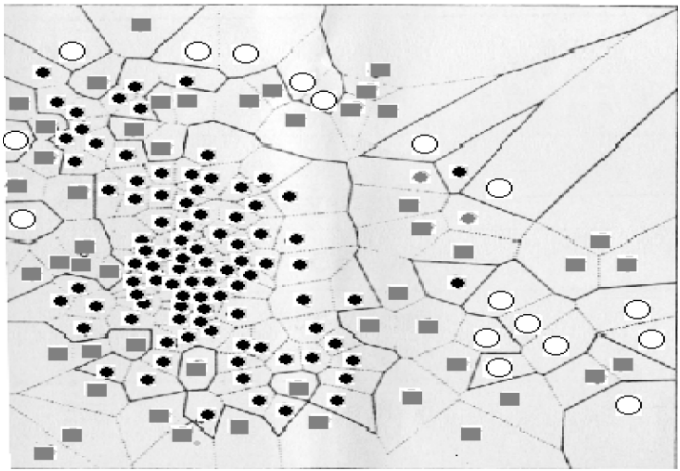

Fig. 1. Annular sorting in a real *Temnothorax* colony (taken from (Franks and Sendova-Franks, 1992), with permission). Black objects are eggs and micro-larvae (youngest), grey objects are medium larvae and white objects are large larvae (oldest).

Wilson *et al.* proposed the first model of "ant-like annular sorting" to simulate the behaviour of *Temnothorax* ants using minimalist robot and computer simulations (Wilson *et al.*, 2004). Three models for annular sorting were presented: "Object clustering using objects of different size", "Extended differential pullback" and "leaky integrator". The first was run exclusively as a computer simulation, since modifying robots to allow them to move objects of different sizes proved to be too complex. Despite this, the computer simulation modelled physical robot behaviour faithfully, preserving the limitations of movement inherent in simple robots, and even going so far as to build in a 1% sensor error that matched the rate seen in the machines.

The first model explored the theory that annular sorting occurs solely due to the different sizes of the objects involved, Wilson *et al.* compared this to the manner in which muesli settles in transit, with smaller objects falling to the bottom, leaving the larger ones on top. The simulation modelled agents who picked up the first object they encountered while unladen and deposited it at the next object they encountered. The results displayed a slight increase in the quality of clusters when there was a greater number of different object sizes; however, clusters tended to form at the edge of the area and were often "inside out", i.e., with larger objects at the centre surrounded by smaller objects. This lead Wilson *et al.* to conclude that merely using objects of different sizes did "not create a sufficient muesli effect".

The second and third models attempted to recreate ant brood clustering behaviour by assigning more complex behaviour to the ants. Wilson *et al.* hypothesised that ants were able to recognise inherent differences in larval growth, and created annular clusters by depositing different objects at different differences away from each other depending on size. Due to the limitations of the robots, this was implemented by having the agents reverse a distance depending on the kind of object carried before depositing it. Initial results with this approach were not good. As a result Wilson *et al.* proposed a third model, calling it the "leaky integrator". This allowed the agents to have an adaptive amount of "pullback" that varied according to how many objects of the same kind had been encountered in the last $n$ seconds. Initial tests with this system produced poor results but results improved when a genetic algorithm was used to select parameter values. Two subsequent models (Hartmann, 2005; Vik, 2005) both use a neural network controller for individual ants, with network weights being evolved using a genetic algorithm. These models have been successfully applied to the problems of clustering and annular sorting of objects (with spatial restrictions imposed, see the later discussion.) Other related work has studied emergent sorting using cellular automata (Scheidler *et al.*, 2006).

## 3. Attraction-Repulsion Algorithm for Annular Sorting

We now propose an alternative algorithm for annular sorting. In contrast to previous work, we focus our attention on the items to be sorted rather than on the agents performing the sorting. Our algorithm is a distributed system in which agents probabilistically pick up or drop items depending on an assessment of the item's "score" (calculated as a function of its current position). Brood items of different sizes are represented by "objects". Agents and objects are spatially distributed at random on a two-dimensional "board" of fixed size.

Each object has a *placement score*; agents move randomly across the board, and when they collide with an object they calculate its placement score. This score is then used to probabilistically determine whether the agent should *pick up* the object and become laden. Laden agents carry objects around the board, and at every time-step they evaluate what placement score the carried object *would have* if it were to be deposited at the current point. This score is then used to probabilistically determine whether the object should be *deposited*.

Given a set of $n$ objects, $L = \{1, 2, \ldots, n\}$, each object, $L_i$, to be sorted has a single attribute: *size*, $s_i$. Size is important, as objects may not overlap. The notion of a *perimeter* is used in the calculation of the object's placement score: When evaluating the placement score of an object, the agent counts how many nearby objects fall within some *minimum* perimeter, $p_{min}$, weighted by object size) (these count towards a penalty, to implement repulsion), and how many fall within some *maximum* perimeter, $p_{max}$, weighted by object size (these count towards a bonus, and implement attraction).

Formally, the placement score is evaluated as follows: we define a Cartesian distance function for two objects $i$ and $j$ as

$$distance(i,j) = \sqrt{(x_i - x_j)^2 + (y_i - y_j)^2}. \tag{1}$$

We then use this to determine the *quality*, $q$, of the placement of $i$ and $j$ (assuming $i \neq j$):

$$q(i,j) = \begin{cases} -60 & \text{if } distance(i,j) \leq (p_{min} * s_i) \\ 0.1 & \text{if } distance(i,j) \leq (p_{max} * s_i) \\ 0 & \text{otherwise.} \end{cases} \quad (2)$$

For a given object, $i$, we may assess its overall placement score by summing over the quality scores for it and all other objects:

$$score(i) = \sum_{j=1}^{n} q(i, L_j). \quad (3)$$

It is important to note that the maximum and minimum perimeters are only calculated for the object whose placement score is currently being evaluated. Agents take no account of whether a placement results in a good or bad score for neighbouring objects, thus small objects with smaller minimum perimeters will frequently be placed quite close to larger objects, resulting in a penalty score for those larger objects. As a result groups of smaller objects will tend to "force out" larger objects. It could be argued that this process implements a stronger version of the size-based muesli analogy proposed in (Wilson *et al.*, 2004). The minimum perimeter of an object acts as a "repulsive" force, whereas the maximum perimeter acts as an "attractive" force, similar to those described in (Couzin *et al.*, 2002). These coupled forces ensure that objects maintain a proportional "exclusion zone", around themselves, whilst ensuring the coherence of a single cluster around some centroid ("centre of gravity"). A precedent for this "attraction-repulsion" model of collective behaviour already exists and has been well-studied in the ecological domain (Gueron *et al.*, 1996; Tien *et al.*, 2004). We may speculate that a plausible biological encoding of perimeter values may use the concentration of some pheromone deposited by ants on objects. However, we have not yet tested this theory, and it is important to note that we are not claiming to have developed a pheromone- *based* algorithm.

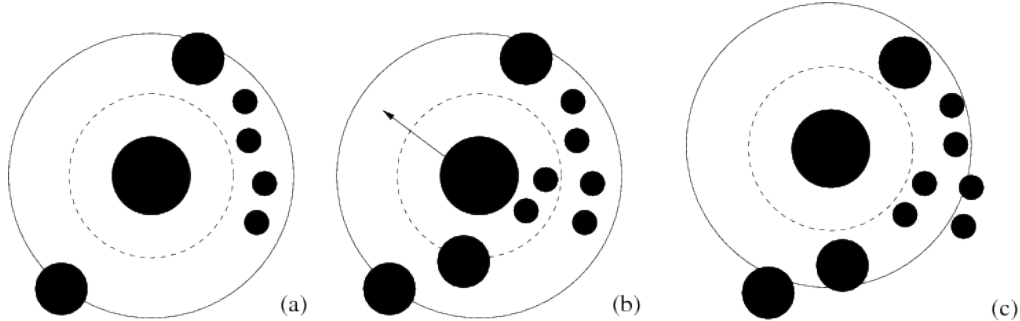

Fig. 2. Depiction of how large objects are forced out of clusters. (a) Good configuration. (b) Placement score of central object drops as a result of three new objects being placed within its minimum perimeter. (c) Object is replaced with a higher placement score.

We now illustrate the underlying principle of our model with an example, depicted in Figure 2. This example shows how larger objects are "forced out" of clusters by their proximity to other objects. The large object in Figure 2 has its minimum perimeter depicted as a dotted line, and its maximum perimeter as a solid line. In Figure 2(a), the object has a high-scoring placement score, as it has no objects within its minimum perimeter (which would attract a penalty), and several objects in its maximum perimeter (attracting a bonus score). However, if several objects are later deposited in the central object's immediate vicinity (Figure 2(b)), its placement score changes for the worse, as these objects contribute a significant penalty. As a result, the next time this object is encountered by an agent it will be carried until it once again has a beneficial placement score (Figure 2(c)), where it will be deposited. As we can see, this calculation of placement scores has the effect of moving larger objects away from clusters of smaller objects, whilst maintaining relative proximity.

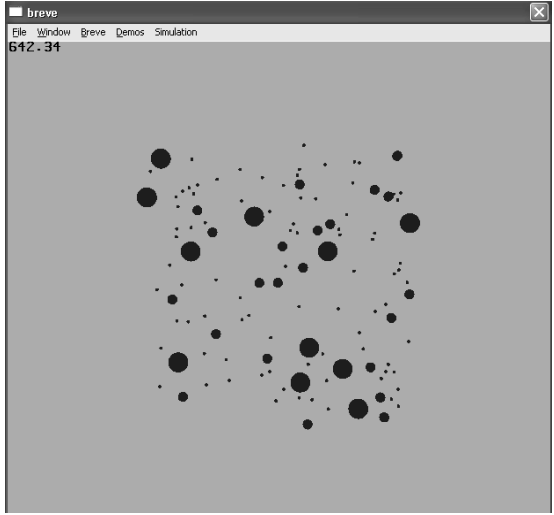

Fig. 3. Poor sorting caused by zero delay ("chaotic" behaviour).

Before embarking on a full implementation of our algorithm, we carried out test runs using a prototype system. This highlighted two, previously unforeseen, effects that we deal with in the full algorithm. Early investigations showed that agents had a tendency to construct good annular clusters, but would then deconstruct them at the outer edge, as there was no termination criterion defined for the algorithm. We initially solved this problem by introducing the notion of "energy"; each agent starts with a fixed amount of energy, represented as an integer value, which is decremented every time the agent picks up an object (the amount of energy lost is a function of the object's size). Once the agent's energy level reaches zero it is removed from the system, and the simulation terminates when there are no more agents left. This prevents the simulation from ending when objects are still being carried, but also has the beneficial effect of smoothly and gradually reducing the overall activity within the system as the simulation approaches termination. The biological validity of this approach is unclear, but it appears to make the placement of outer objects much more realistic.

Another problem that was highlighted by the prototype algorithm was that of "chaos"; agents would frequently deposit an object and then immediately pick it up again, as they were still in contact with it. Conversely, agents would also frequently deposit objects after taking a single "step", particularly if the placement score was moderately good. This behaviour creates clusters in which overall placement was quite poor, as large objects were rarely removed from the centre of the cluster (Figure 3). In order to deal with this problem, a "cooling down" delay period was introduced; agents must wait a set number of steps after colliding with or picking up an object before they may carry out any further placement calculations. This appeared to solve the problem; different (non-zero) values for the delay variable impacted only on the run-time of the simulation, and did not affect the quality of clusters generated. A delay value of 4 was chosen for what follows.

### 3.1. *The algorithm*

We now describe in detail our algorithm for annular sorting. Ants are modelled by *agents*, each of which has the following attributes:

- Location (application representation)
- Laden (true/false)
- Object (application representation of object carried if Laden == true)
- Energy (integer)
- Delay (integer)

Objects are modelled as spheres, and have a single attribute, from which their minimum and maximum perimeters are calculated:

- Size (real from the set $\{0.5, 1.5, 3\}$, corresponding to small, medium or large)

The following constants are defined to deal with objects (all distances are measured from the edge of objects, rather than from their centre):

- BONUS (0.1)
- PENALTY (-60)
- $p_{max}$ (4.0) (to calculate maximum perimeter)
- $p_{min}$ (0.4) (to calculate minimum perimeter)

There are $n$ objects and $m$ agents initially distributed at random on a two-dimensional "board" of fixed size. The number of agents and numbers of objects of each size may be specified in advance. Agents may move over other agents or over objects; this is in contrast to previous work modelling robotic agents, where inherent spatial restrictions exist. We impose no such limitations, and discuss in a later section the implications for comparison of results. Movement may occur continuously in any direction on the Cartesian plane; we do not impose a discrete, cell-based "neighbourhood". The algorithm is depicted in flowchart form in Figure 4. The pseudo-code expression of the algorithm is as follows:

```
while (agents exist)
    for all agents (A₁,…Aᵢ,…Aₘ) do
       // Remove agent if "dead"
       if (Aᵢ.Energy == 0)
          remove Aᵢ from system
          break // i.e., go to next agent, if possible
       end if

          // Check delay
          if (Aᵢ.Delay > 0)
             Aᵢ.Delay = Aᵢ.Delay-1
          else
             // Unladen agent collides with object
             if (Aᵢ.Laden == false and (Aᵢ.Location == some Oᵢ.Location))
                Aᵢ.Delay = 4
                Score = CalculatePlacement(Oᵢ)
                Probability = random(0…1)
                if (Score < Probability)
                   Aᵢ.Object = Oᵢ
                   Aᵢ.Laden = true
                   Aᵢ.Energy = Aᵢ.Energy-Oᵢ.Size
                end if
             end if

             // Laden agent in free space
             if (Aᵢ.Laden == true and (Aᵢ.Location == empty))
                Score = CalculatePlacement(Aᵢ.Object)
                Probability = random(0…1)
                if (Score > Probability)
                   Place Aᵢ.Object at Aᵢ.Location
                   Aᵢ.Delay = 4
                   Aᵢ.Laden = false
                end if
             end if
          endif
       Move Aᵢ to a randomly selected adjacent location
    end for
end while
```

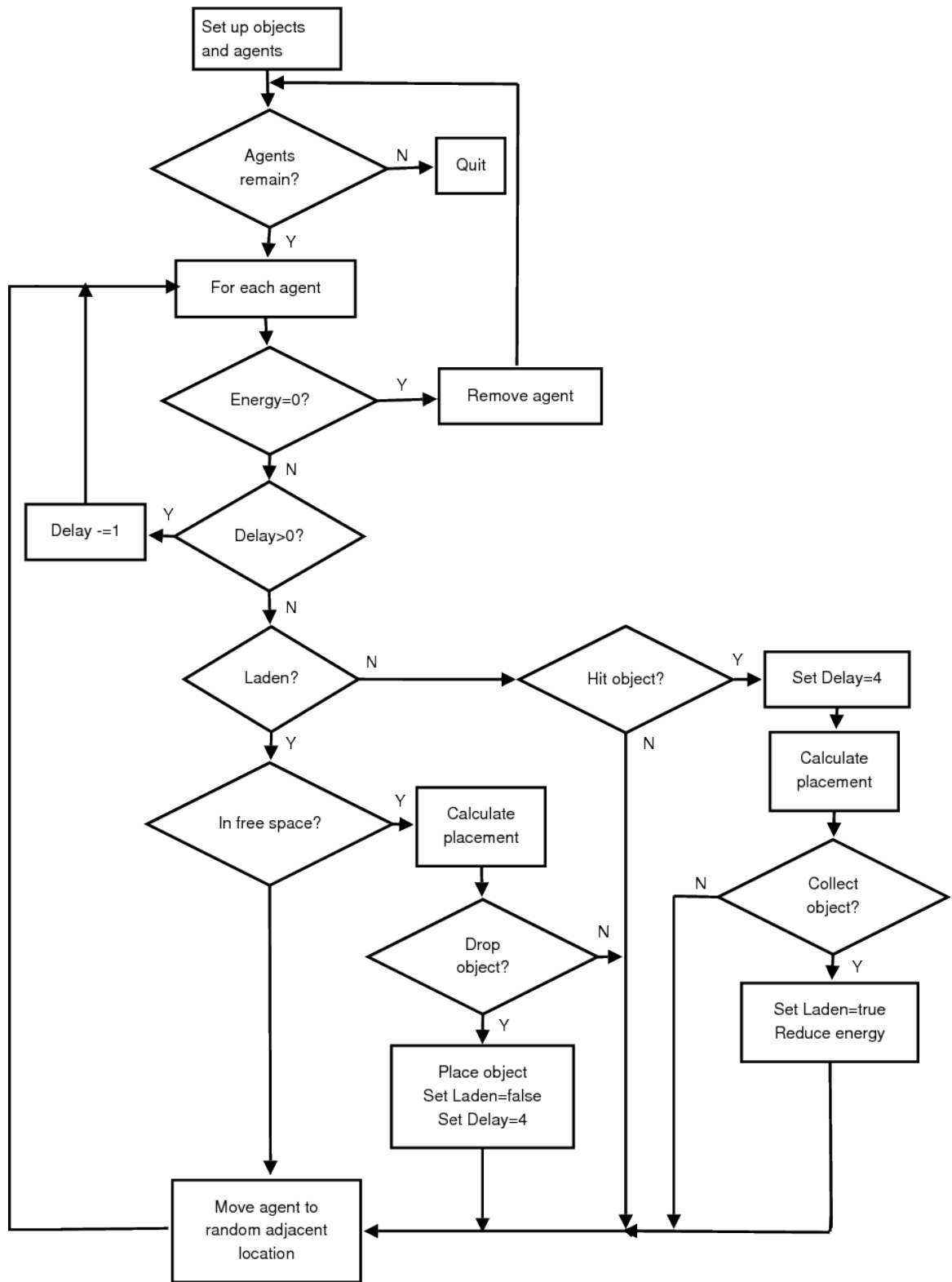
Fig. 4. Attraction-repulsion algorithm flowchart.

We now describe the CalculatePlacement() function:

```
CalculatePlacement(Object)
   Score = 0
   for each neighbouring Object L_i (where Object ≠L_i)
   // "within"calculates Cartesian distance
      if (L_i.Location within (P_min * Object.Size))
         Score = Score + PENALTY
      else
      if (L_i.Location within (P_max * Object.Size))
         Score = Score + BONUS
      end if
   end for
return (Score)
```

### 3.2. *Implementation*

The pseudo-code above was implemented using the Breve[2] multi-agent system environment (Klein, 2002). This package facilitates the simulation of decentralised systems, in a similar fashion to the well-known Swarm[3] system. The benefits of using such a framework are derived from the fact that it automatically handles issues such as collision detection, construction of object neighbourhoods, and the user interface. This allows the programmer to concentrate effort on the important aspects of the model's implementation, rather than on "house keeping" issues. All simulations were run using Breve version glut/2.6 (Sep. 8 2007) on an PC (AMD Athlon™ processor, 768MB RAM) under Ubuntu Linux 7.10, although the code runs equally well on both Windows and Apple Macintosh machines. An additional Java application was written to deal with post-processing of object coordinates.

### 4. Quality Metrics

In order to assess the quality of sorted structures, we apply three performance metrics: *separation*, *shape*, and *radial displacement*, as defined in previous work (Wilson *et al.*, 2004). Separation and shape are expressed as a percentage, with a value of 100% being interpreted as ideal. Separation measures the degree to which objects of similar size are kept apart from objects of differing size (i.e., the degree of "segregation"). The distance to the structure centroid is calculated for each object, and the upper and lower quartiles computed for each object type. We then perform three individual counts:

1. $N_s$ = the number of small objects that have a distance to the centroid *greater* than the lower quartile range of either the medium or the large objects,
2. $N_l$ = the number of large objects that have a distance to the centroid *less* than the upper quartile range of either the medium or the small objects,
3. $N_m$ = the number of medium objects that have a distance to the centroid *greater* than the lower quartile range of the large objects, plus the number of medium objects that have a distance to the centroid *less* than the upper quartile range of the small objects. This count is then divided by two to prevent bias.

Separation is therefore expressed for $n$ objects as

$$Se = 100 * \left(1 - \frac{N_s + N_l + \frac{N_m}{2}}{n}\right). \tag{4}$$

The shape metric is used to assess the "circularity" of the structure generated, with the ideal structure being a compact core of small objects, surrounded by perfectly circular successive bands of larger objects. This metric is calculated in two stages. We first calculate $F_s$, the fraction of small objects located in the central cluster. We achieve this by constructing a graph, with each vertex representing a small object, and an edge connecting two vertices if the corresponding objects are within 2.5 spatial units of one another. We then divide the size of the largest connected component of this graph by the total number of small objects. The second stage of the shape calculation involves finding the deviation from some common radius for each object size, since each object would ideally lie on the same radius as every other object of that size. For the medium and large objects, we first calculate the common radius by taking the mean radial distance from the centroid ($r_m$ and $r_l$ for medium and large objects). We then calculate the absolute deviation from this radius by summing the Cartesian distances to each object of a particular type from this radius ($d_m$ and $d_l$ for medium and large objects). We then calculate this as a percentage by placing this sum deviation between zero and one common radius. Shape is therefore expressed as (cluster fraction + sum of shape performances for medium and large objects)/number of object sizes:

$$Sh = (100 F_s + (100 * (1 - (d_m/n_m * r_m))) \\ + (100 * 1 - ((d_l/n_l * r_l))))/3 \quad (5)$$

---
[2]Available at http://www.spiderland.org/breve/
[3]http://www.swarm.org

Radial displacement is used to measure the "compactness" of a structure, and yields a distribution of distances from the centroid for each object type. Previous studies (Wilson *et al.*, 2004; Hartmann, 2005) provide precise formulae for the calculation of compactness for a given structure, but this is difficult for our model. Earlier work used objects of uniform size, which makes the task of calculating an optimal "packing" relatively straightforward. Here, however, we use objects of non-uniform size, and little work has been done on packing collections of such objects.

## 5. Results

We should note that it is difficult to draw direct comparisons between our results and those of (Wilson *et al.*, 2004), as their model enforces strict spatial constraints on the movement of agents and objects. In addition, (Hartmann, 2005) presents results only in the context of genetic algorithm fitness evaluations, with no individual breakdowns for each metric, so direct comparisons are again difficult (although this paper does use the same separation and shape algorithms as the those used by (Wilson *et al.*, 2004) and ourselves). Nonetheless, the metrics provide a useful standardised framework for performance analysis.

Three sets of initial trials were carried out to investigate the algorithm's basic performance, given the following initial configurations:

1. Equal numbers of each object, randomly distributed (as in (Wilson *et al.*, 2004; Hartmann, 2005)),
2. Unequal numbers of each object, randomly distributed (as observed in nature (Franks and Sendova Franks, 1992)),
3. Pre-sorted, equally-sized clusters of objects.

### 5.1. *Equal object numbers*

The first set of experiments replicated the initial conditions described in (Wilson *et al.*, 2004): 15 objects of each type, randomly distributed across the surface, with 6 agents. The average separation and shape scores for 50 initial configurations were 11.85% and 48.21% respectively. The results obtained are depicted in Table I, with the best figures obtained highlighted in bold. The radial displacement distributions and a typical final pattern are depicted in Figure 5.

The figures of 79.52% and 70.88% for separation and shape respectively compare well with the figures of 59% and 68.5% obtained by the leaky integrator of (Wilson *et al.*, 2004) (noting that their simulation includes extra spatial constraints and uses objects of uniform size, whilst ours has no such constraints but handles objects of different sizes).

Table I. Results for 15 objects of each type with 6 agents, averaged over 50 runs for each energy value.

| Energy | Separation | Shape |
|---|---|---|
| 250 | 51.38 | 56.49 |
| 500 | 61.71 | 60.17 |
| 750 | 71.09 | 64.65 |
| 1000 | 77.19 | 68.21 |
| 1250 | 77.14 | 68.96 |
| **1500** | **79.52** | **70.88** |
| 1750 | 79.0 | 70.86 |

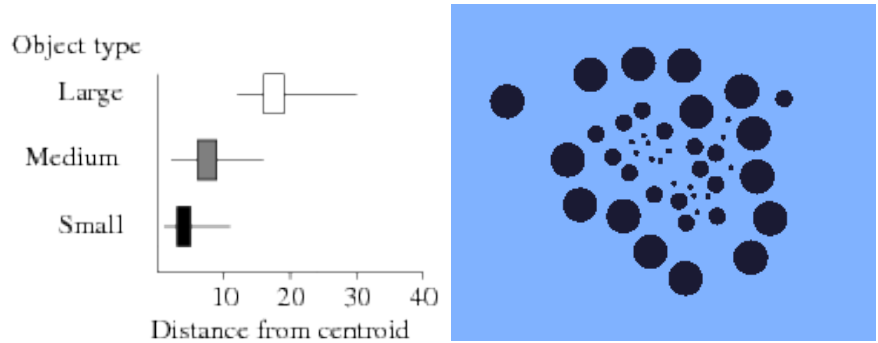

Fig. 5. Equal numbers of objects. (Left) Radial displacement. Minimum and maximum distances are represented by the lines, and interquartile ranges by the boxes. (Right) Typical final structure obtained.

We also carried out a set of 50 control trials, where the CalculatePlacement function was replaced by a "coin flip" (that is, the agents would pick up or drop the object, on average, 50% of the time). In the case of equal object numbers, with an energy value of 1500, we obtained average figures of 13.24% for separation and 41.77% for shape.

### 5.2. *Unequal object numbers*

The aim of the second set of experiments was the assess the algorithm against the type of configuration observed in actual *Temnothorax* nests; that is, where there are many more small brood items than large (i.e., older) items (Franks and Sendova Franks, 1992) (see Figure 1). In these experiments, we randomly distributed 40 small objects, 20 medium objects and 10 large objects. In order to retain the agent-to-object ratio used in the previous set of experiments, we used 10 agents in this set. The average separation and shape scores for 50 initial configurations were 17.16% and 48.25% respectively. The results obtained are depicted in Table II, with the best figures obtained highlighted in bold. The radial displacement distributions and a typical final pattern are depicted in Figure 6.

Table II. Results for |Small|=40, |Medium|=20, |Large|=10 with 10 agents, averaged over 50 runs for each energy value.

| Energy | Separation | Shape |
|---|---|---|
| 250 | 76.86 | 75.10 |
| 500 | 87.00 | 81.30 |
| **750** | **93.47** | **84.62** |
| 1000 | 93.04 | 84.37 |

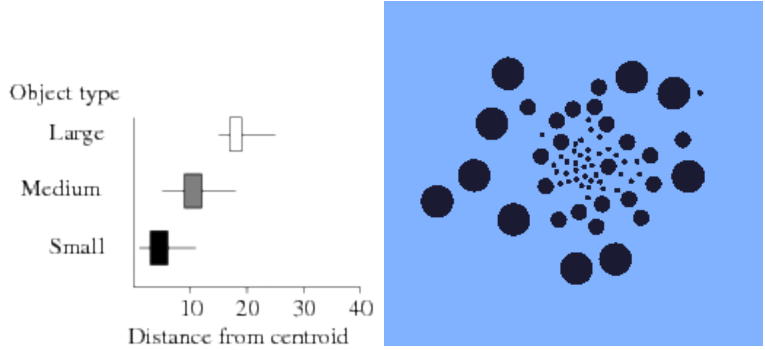

Fig. 6. Unequal numbers of objects. (Left) Radial displacement. (Right) Typical final structure obtained.

In the control trials, with energy set to 750, we obtained average figures of 16.84% for separation and 40.36% for shape. The algorithm clearly performs best when applied to distributions of objects that roughly match those observed in nature. The high separation score of 93.04% is in general partly due to the observed creation of a large, densely-packed core of small objects at the centre of the structure. Once built, this core is rarely disturbed by the agents, and sorting only occurs in the outer bands.

### 5.3. *Pre-sorted clusters*

The aim of the second set of experiments was the assess the algorithm's ability to perform annular sorting of objects that were pre-sorted into piles. We created three piles, each one consisting of 15 objects of a particular size randomly clustered around a fixed point (Figure 7). As in the first experiment, 6 agents were used. The separation and shape scores for initial configurations are clearly meaningless in this context, so we omit them here. The results obtained are depicted in Table III, with the best figures obtained highlighted in bold. The radial displacement distributions and a typical final pattern are depicted in Figure 8.

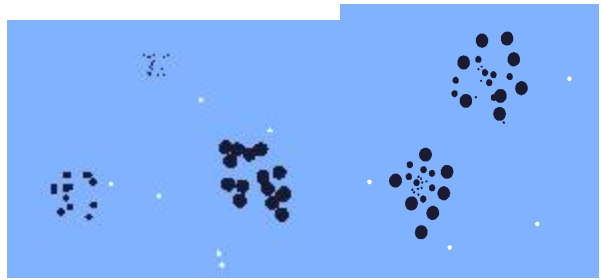

Fig. 7. Pre-sorted objects. (Left) Initial configuration. (Right) Intermediate stage in sorting.

Table III.  Results for 15 pre-sorted objects of each type with 6 agents, averaged over 50 runs for each energy value.

| *Energy* | *Separation* | *Shape* |
|---|---|---|
| 250 | 41.71 | 55.46 |
| 500 | 64.14 | 62.82 |
| 750 | 75.19 | 67.02 |
| **1000** | **77.23** | **68.95** |
| 1250 | 76.67 | 67.87 |

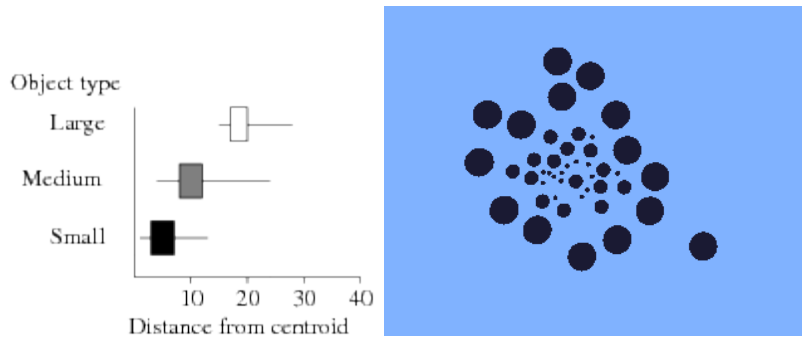

Fig. 8. Pre-sorted objects. (Left) Radial displacement. (Right) Typical final structure obtained.

Our studies show that the algorithm is able to convert a pre-sorted configuration into one that is sorted in an annular fashion. Given sufficient energy, there is little difference in performance in sorting either pre-sorted or randomly distributed configurations. Observation of the algorithm shows that, in general, the agents form two clusters of roughly equal size and composition (Figure 7). These are gradually merged into a single structure which is then refined in terms of shape and separation. It is important to note that no modifications (either to the model code or to the parameters) were necessary in order for these results to obtained. This suggests that the model is robust and capable of dealing with a variety of initial configurations.

### 5.4. *Parametric analysis*

After establishing the effectiveness of the basic algorithm, we turned our attention to investigating the effect of altering various parameters. In what follows, we performed 50 runs for each parameter setting and took the average performance value.

### 5.4.1. *Energy allocation*

We first investigated the effect of changing the amount of energy allocated to each agent, the idea being to establish the optimal amount, given that termination of the algorithm only occurs when every agent's energy is exhausted. In these sets of experiments, we used one agent per object.

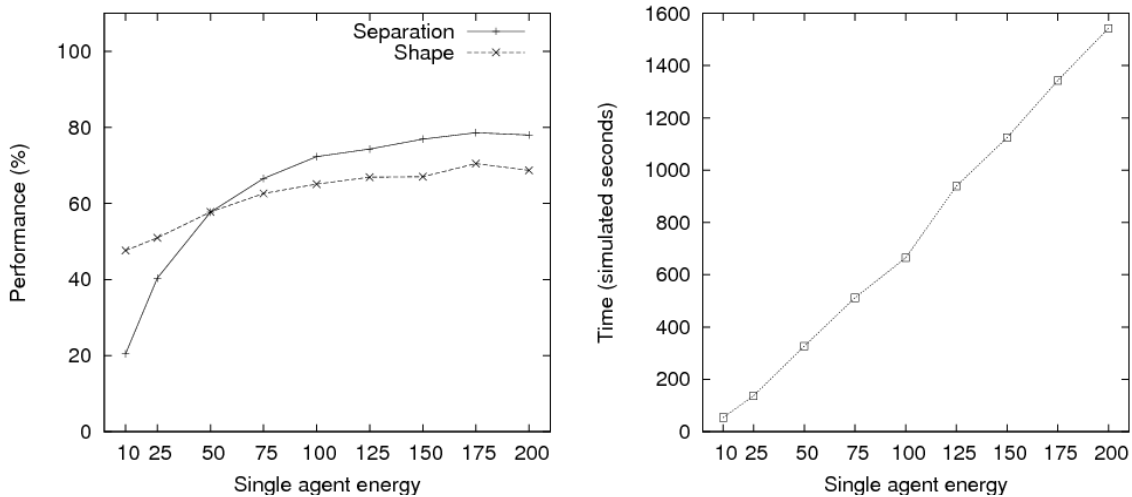

Fig. 9. Varying energy allocated to each agent; uniform distribution of object sizes.

In the case of uniform object numbers (Figure 9), we began by giving each of the 45 agents 10 units of energy, and then gradually increased this amount up to a maximum of 200. The previous experiments suggested that no performance benefit could accrue beyond this point (9000/45=200), which was confirmed by this set of trials. Both performance curves began to flatten at around 100, and no increase was seen after 200 units. Run time increased linearly with increases in energy.

In the mixed case (Figure 10), we had 70 agents, with a maximum net energy of 7500 units (from Table II). Energy was set to and increased by the same proportional amount as in the uniform case; the uniform trials started with energy set to $10/45 = 0.22$, so the mixed trials started with energy of $70 * 0.22 = 15$. The next uniform energy level was 25, and $25/45 = 0.56$, so the next mixed energy level was $70 * 0.56 = 39$, and so on.

The uniform situation (Figure 9) required rather more energy to achieve stability than the the mixed situation (Figure 10); we believe that this is due again to the formation, in the mixed case, of a core of small objects which are then rarely disturbed. Again, we observed a linear relationship between energy and run time.

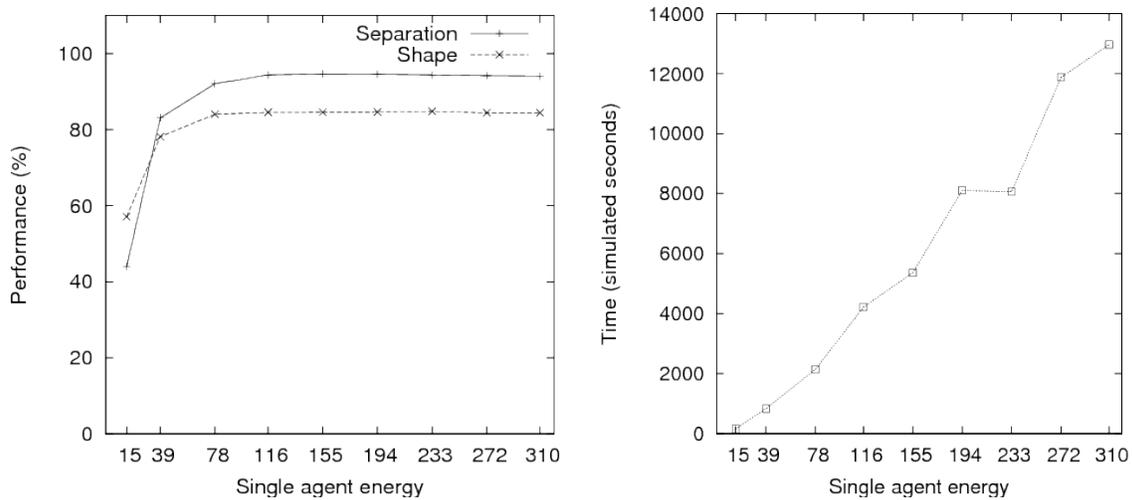

Fig. 10. Varying energy allocated to each agent; mixed distribution of object sizes.

### 5.4.2. *Ratio of agents to objects*

We then examined the effect of the ratio of agents to objects, the idea being to establish the point at which collective (as opposed to individual) computation becomes effective. For each set of such trials, we established, from the previous experiments the optimal net energy in the system, and then *distributed* this over a varying number of agents. For example, we already established that the optimal system energy in the uniform case was $45 * 200 = 9000$ units, so for the first uniform trial we used a single agent with energy equal to 9000. Beginning with a single agent, we gradually increased agent numbers (reducing the unit energy accordingly) until agents outnumbered objects by a factor of 50%.

Clearly, from Figures 11 and 12, the ratio of agents to objects has little effect on the overall *quality* of the solutions generated. Both sets of performance metrics are in line with those previously observed. However, the average *duration* of a run varied dramatically, with small numbers of agents yielding large run times (remembering that runs are terminated by the exhaustion of energy). In both cases (uniform (Figure 11) and mixed (Figure 12) distribution of object numbers), average run time stabilises when the number of agents is roughly half that of the objects. After this point, adding extra agents appears to have no significant effect on reducing run time.

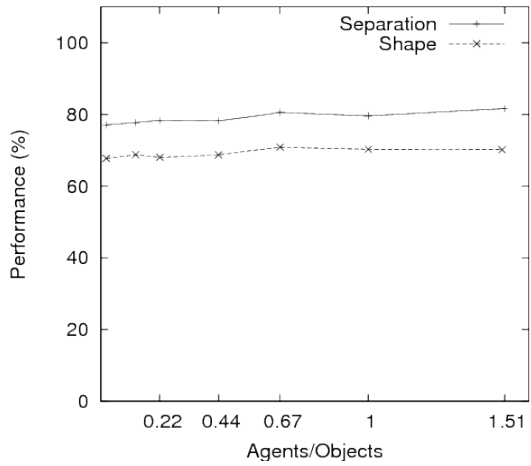 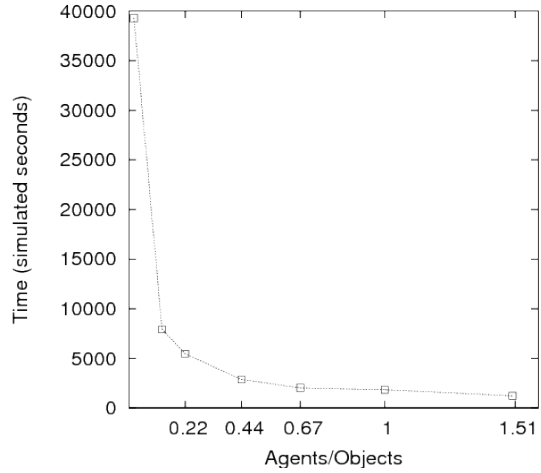

Fig. 11. Varying ratio of agents to objects; uniform distribution of object sizes.

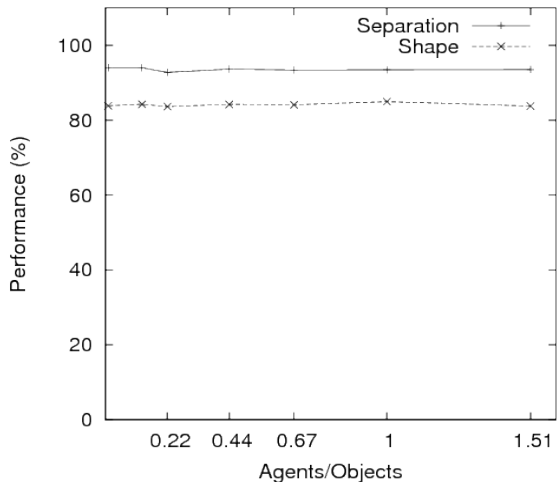 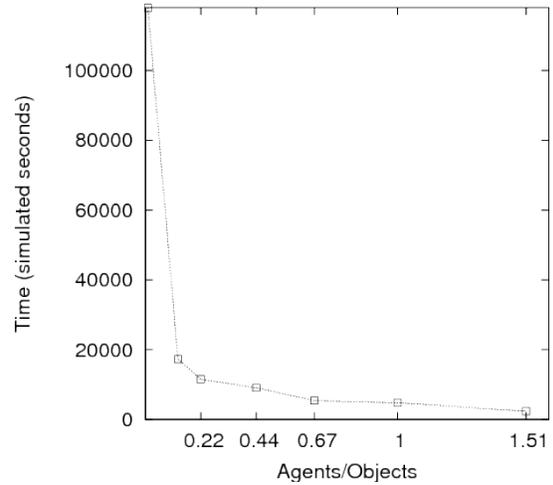

Fig 12. Varying ratio of agents to objects; mixed distribution of object sizes.

### 5.4.3 *Fixed energy penalty*

This set of experiments concerned the biological realism of forcing each agent to expend an amount of energy proportional to the size of the object carried. We performed a set of control trials, where energy is removed as described in the original algorithm, and then ran a series of trials where the energy penalty for moving an object was *fixed*, regardless of its size. The agent numbers and their initial energy values were determined using the results obtained from the previous sets of experiments. In the uniform case (Table IV), we ran with $45 * 0.44 = 20$ agents, each given $4.44 * 20 = 88$ initial units of energy. In the mixed case (Table V), we had $70 * 0.67 = 47$ agents, each with $1.67 * 47 = 79$ units of energy.

Table IV. Uniform object distribution, varying energy penalty.

| Penalty | Separation | Shape | Time |
|---|---|---|---|
| Control | 72.71 | 65.07 | 1196 |
| 0.5 | 72.1 | 64.24 | 1326 |
| 1.5 | 53.95 | 57.27 | 509 |
| 3 | 46.81 | 53.54 | 278 |

Table V. Mixed object distribution, varying energy penalty.

| Penalty | Separation | Shape | Time |
|---|---|---|---|
| Control | 94.57 | 85.32 | 5154 |
| 0.5 | 92.66 | 83.55 | 4880 |
| 1.5 | 87.76 | 81.11 | 1633 |
| 3 | 78.86 | 74.35 | 903 |

The results (Tables IV and V) suggested that a size-dependent penalty is *moderately* beneficial. A large fixed penalty led to premature convergence of the algorithm, as the system energy was expended before the agents have had a chance to construct a good configuration. Conversely, a small fixed penalty did not offer any improvement over the control (apart from a small reduction in run time in the mixed case).

### 5.5. *Convergence analysis*

In the final set of experiments, we performed some trials without the use of energy, choosing instead to terminate the algorithm after a fixed number of "steps". The aim here was to investigate the convergence behaviour of the algorithm for different initial configurations, and to establish (based on earlier discussions (Melhuish, 2005)) whether or not the use of energy provided a satisfactory termination method.

For each initial configuration type, we first varied the number of agents, and investigated the relationship between population size and convergence of the task towards "completion" (in terms of separation and shape performance). For each trial we define a *step* as the execution of one agent's instructions, assessed the quality of the configuration every 25,000 steps, and terminated the run after 1,000,000 steps. As in previous experiments, results were averaged over 50 trials.

The results obtained are depicted in Figures 13 and 14. Based on these results, we then investigated the impact of the choice of termination mechanism (energy or steps) on the *real* elapsed run-time of the algorithm. In each case, we ran 50 trials, one set using energy termination, and the other terminated after a fixed number of steps.

In the first experiment, using 15 objects of each type, we used $45 * 0.57 = 25$ agents for each set of trials, and in the energy trials each agent was given $9000/25 = 360$ units. In order to ascertain the termination step we first ran the energy trials, and then used the average separation and shape scores in conjunction with the plots in Figure 13 to estimate its value at 700,000.

In the second experiment, using the mixed set of objects, we used $70 * 0.57 = 40$ agents for each set of trials, and in the energy trials each agent was given $7500/40 = 188$ units. As before, we ran the energy trials first, and then used the plots from Figure 14 to estimate a termination step of 900,000.

In both cases, the use of energy as the termination mechanisn led to high-quality final configurations, but the use of steps facilitated comparable results in a shorter period of time (Tables VI and VII). Future work will consider the scalability of the algorithm, and attempt to derive general guidelines concerning the choice of termination conditions.

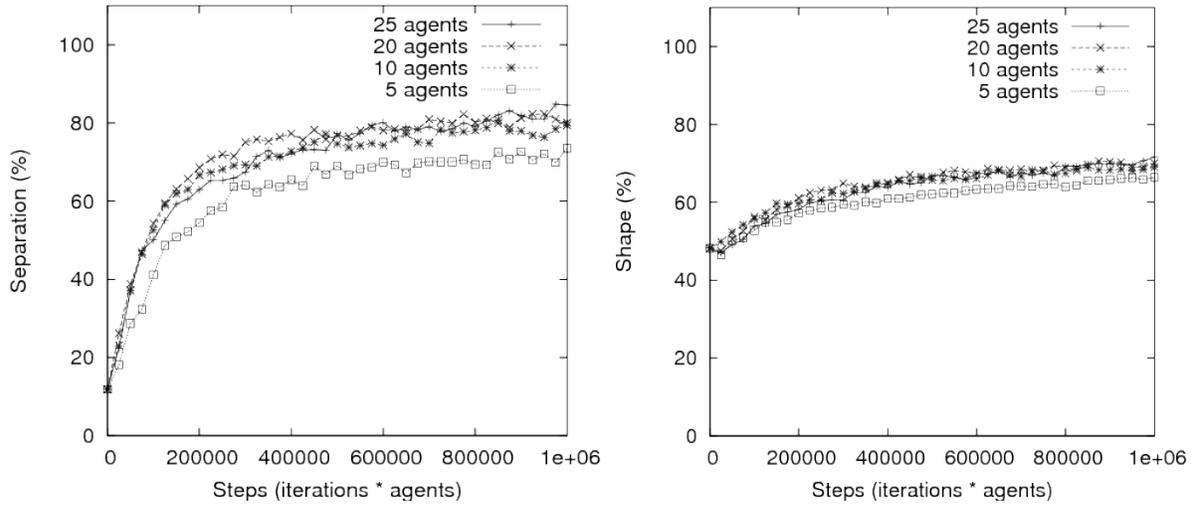

Fig. 13. Convergence behaviour; uniform distribution of object sizes.

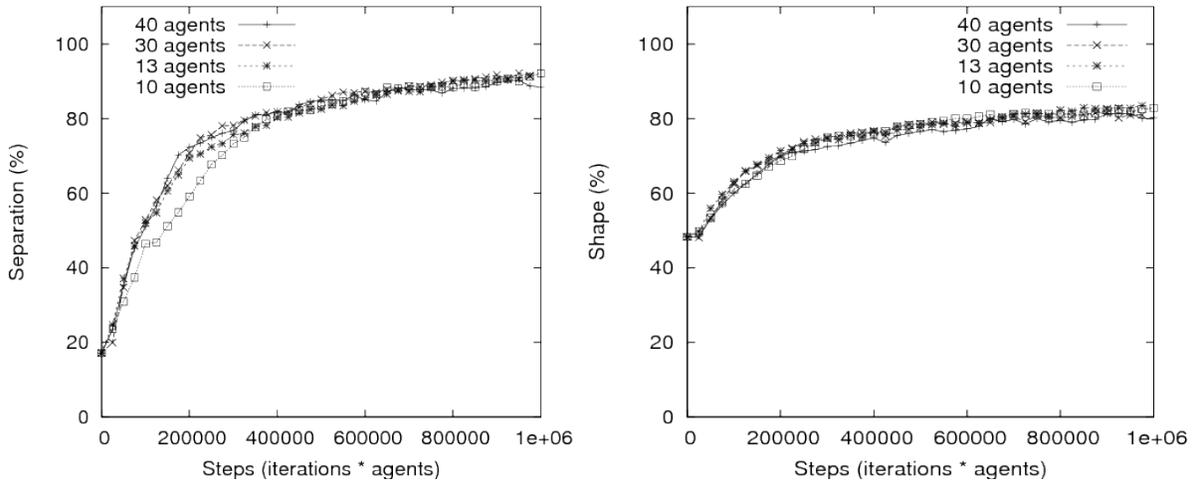

Fig. 14. Convergence behaviour; mixed distribution of object sizes.

Table VI. Quality and elapsed real time comparison for two termination methods; uniform distribution of objects.

| Termination | Separation | Shape | Time (s) |
| --- | --- | --- | --- |
| Energy | 81.00 | 71.07 | 101.83 |
| Steps | 78.05 | 68.2 | 88.80 |

Table VII. Quality and elapsed real time comparison for two termination methods; mixed distribution of objects.

| Termination | Separation | Shape | Time (s) |
| --- | --- | --- | --- |
| Energy | 91.4 | 83.3 | 378.25 |
| Steps | 89.7 | 80.53 | 190.77 |

### 5.6. *Discussion*

Although we present no formal theoretical analysis of our model, recent related work (Gaubert *et al.*, 2007) supports our observation that a key component in the construction of annular structures is that of the *perimeters* (referred to by Gaubert *et al.* as the "virtual sizes" of objects). They argue that "...we believe that we have highlighted two of the major causes of the concentric annuli formation in *Leptothorax unifasciatus* ant colonies, that is: almost minimization of the average of squared-distances between free larvae, coupled with great differences between virtual sizes of brood items." The first cause relates to a notion of "attraction" (similar to ours), in which annular sorting arises as a result of *local* minimization processes. The second cause may also be observed in our own studies, where the "virtual sizes" of small, medium and large brood items are 2, 6 and 12 units respectively (that is, substantially different). A key finding of (Gaubert *et al.*, 2007) is that swarm-based activity is not *necessary* for the formation of annular structures, which we confirm in Section 5.4.2. However, we find that the *time required* to build an annular structure is indeed greatly affected by the ratio of agents to objects, which may well inform subsequent practical applications of the model.

### 6. Conclusions and Future Work

This study has demonstrated a swarm-based model that can consistently sort objects into annular clusters. There are real world examples where this might be useful, particularly when one has objects that may change over time and require different amounts of separation. Our algorithm uses only stochastic agent behaviour and a brood item "attraction-repulsion" mechanism. However, some of the aspects that set this study apart from others such as (Wilson *et al.*, 2004) also make it harder to create a physical implementation (perhaps using mobile robots). In particular the model relies on the fact that agents are able to pull poorly placed objects out of the centre of a cluster without disturbing other objects or risking collision, something that is hard to do with simple robotics. These difficulties will become less significant as the state of the art advances. In theoretical terms, more work is required on analysis of our algorithm's convergence properties; similar work in related fields such as particle swarm optimization has generated good results, so we are hopeful that the algorithm will soon be solidly grounded in theory to augment existing empirical work. Future work will study this in further detail, as well as the broader biological significance of the "attraction-repulsion" mechanism. We have a particular interest in modelling biological systems at levels both above and below that of individual organisms, and the notion of attraction-repulsion has clear significance for both molecular and cellular self-assembly and related macro-scale biological phenomena, such as the formation of biofilms or spatio-temporal patterns in response to stress.

### Acknowledgements

The authors are grateful to Jon Klein, Chris Melhuish and Ana Sendova-Franks for useful advice and suggestions, and to Adam Sampson and Paul Andrews for highlighting errors in an earlier version of the pseudo-code. We also thank several anonymous reviewers for their valuable comments.

### References

Amos, M. and Don, O. (2007) An ant-based algorithm for annular sorting. In *Proceedings of the 2007 IEEE Congress on Evolutionary Computation (CEC'07), Singapore, September 25-28, 2007*, pages 142-148. IEEE Press.

Camazine, S., Deneuborg, J-L., Franks, N.R., Sneyd, J., Theraulaz, G. and Bonabeau, E. (2001) *Self-organization in biological systems*. Princeton University Press.

Couzin, I.D., Krause, J., James, R., Ruxton, G.D. and Franks, N.R. (2002). Collective memory and spatial sorting in animal groups. *Journal of Theoretical Biology* **218**(1): 1-11.

Deneubourg, J-L., Goss, S., Franks, N.R., Sendova-Franks, A.B., Detrain, C. and Chretien, L. (1991). The dynamics of collective sorting: Robot-like ants and ant-like robots. In Meyer, J-A. and Wilson, S., editors, *Proceedings of the First International Conference on Simulation of Adaptive Behaviour: From Animals to Animats 1*, pages 356-365. The MIT Press.

Dorigo, M., Di Caro, G. and Gambardella, L.M. (1999). Ant algorithms for discrete optimization. *Artificial Life* **5**(2): 137-172.

Dorigo, M. and Stützle, T. (2004). *Ant colony optimization*. The MIT Press.
Franks, N.R. and Sendova-Franks, A.B. (1992) Brood sorting by ants: distributing the workload over the work-surface. *Behavioural Ecology and Sociobiology* **30** (2): 109-123.
Gaubert, L., Redou, P., Harrouet, F. and Tisseau, J. (2007). A first mathematical model of brood sorting by ants: Functional self-organization without swarm-intelligence. *Ecological Complexity* **4** (4): 234-241.
Gueron, S., Levin, S.A. and Rubenstein, D.I. (1996). The dynamics of herds: from individuals to aggregations. *Journal of Theoretical Biology* **182** (1): 85-98.
Handl, J., Knowles, J. and Dorigo, M. (2003). Ant-based clustering: a comparative study of its relative performance with respect to *k*-means, average link and 1d-som. Technical Report TR/IRIDIA/2003-24, IRIDIA, Universite Libre de Bruxelles, July 2003. http://www.handl.julia.de
Hartmann, V. (2005). Evolving agent swarms for clustering and sorting. In *Proceedings of the 2005 conference on Genetic and Evolutionary Computation (GECCO 05)*, pages 217-224. ACM Press.
Klein, J. (2002). breve: a 3D simulation environment for the simulation of decentralized systems and artificial life. In *Proceedings of Artificial Life VIII, the 8th International Conference on the Simulation and Synthesis of Living Systems*, pages 329-334. The MIT Press, 2002.
Melhuish, C. (2005). Personal communication.
Okubo, A. (2001). *Diffusion and ecological problems: modern perspectives*. Springer.
Parrish, J.K. and Hamner, W.M., editors (1997). *Animal groups in three dimensions*. Cambridge University Press.
Scheidler, A., Merkle, D. and Middendorf, M. (2006). Emergent sorting patterns and individual differences of randomly moving ant like agents. In Artmann, S. and Dittrich, P., editors, *Explorations in the complexity of possible life: abstracting and synthesizing the principles of living systems*, pages 105-116. IOS Press.
Scholes, S., Wilson, M., Sendova-Franks, A.B. and Melhuish, C. (2004). Comparisons in evolution and engineering: the collective intelligence of sorting. *Adaptive Behavior* **12**(3-4): 147-159.
Tien, J.H., Levin, S.A. and Rubenstein, D.I. (2004) Dynamics of fish shoals: identifying key decision rules. *Evolutionary Ecology Research* **6**(4): 555-565.
Vik, A.H. (2005). Evolving annular sorting in ant-like agents. In *Proceedings of the European Conference on Artificial Life (LNAI 3630)*, pages 594-603. Springer-Verlag.
Wilson, M., Melhuish, C., Sendova-Franks, A.B. and Scholes, S. (2004). Algorithms for building annular structures with minimalist robots inspired by brood sorting in ant colonies. *Autonomous Robots* **17**(2): 115-136.

**About the authors**

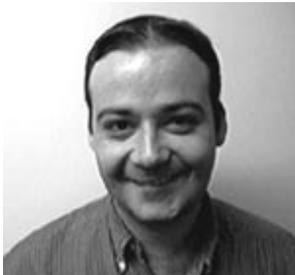 MARTYN AMOS received his B.Sc. (Hons) degree in Computer Science from Coventry University, UK in 1993, and his Ph.D. in DNA Computation from the University of Warwick, UK in 1997. He then held a Leverhulme Special Research Fellowship, before taking up Lectureships in Bioinformatics, first at the University of Liverpool, UK and then at the University of Exeter, UK. He is currently a Senior Lecturer and Leader of the Novel Computation Group in the Department of Computing and Mathematics, Manchester Metropolitan University, UK. He is broadly interested in the development and analysis of computational methods based on natural or self-organizing systems, with particular reference to molecular and cellular computing and population-based approaches. Email:M.Amos@mmu.ac.uk.

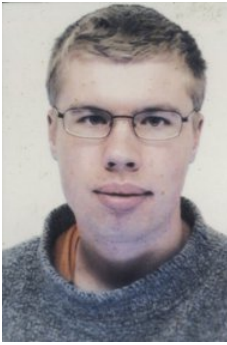 OLIVER DON received his B.Sc. (Hons) degree in Computer Science from the University of Exeter, UK in 2005, and his M.Sc. in Bioinformatics from the University of Edinburgh, UK in 2006. He now works on Location Based Services at Symbian. Email: oliver.don@symbian.com.